\documentclass[letterpaper]{article}
\usepackage{uai2013}
\usepackage{natbib}

\usepackage{xr}
\usepackage{amssymb,amsmath,array}
\usepackage{amsthm}
\usepackage{graphicx} 
\usepackage{caption}
\usepackage{subfig}
\usepackage{algorithm,algorithmic}
\usepackage{cleveref}

\newcommand{\rightpath}{\Rightarrow}
\newcommand{\leftpath}{\Leftarrow}
\newcommand{\confounded}{\Leftrightarrow}
\newcommand{\nopath}{\nLeftrightarrow}

\newtheorem{definition}{Definition}

\title{Scoring and Searching over Bayesian Networks with Causal and Associative Priors}

\author{{\bf Giorgos Borboudakis} \\
Comp. Sci. Dept., University of Crete \\
Institute of Computer Science, FORTH \\
\And
{\bf Ioannis Tsamardinos}  \\
Comp. Sci. Dept., University of Crete \\
Institute of Computer Science, FORTH \\
}

\begin{document}

\maketitle

\begin{abstract}

A significant theoretical advantage of search-and-score methods for learning Bayesian Networks is that they can accept informative prior beliefs for each possible network, thus complementing the data. 
In this paper, a method is presented for assigning priors based on beliefs on the presence or absence of certain paths in the true network. 
Such beliefs correspond to knowledge about the possible causal and associative relations between pairs of variables. 
This type of knowledge naturally arises from prior experimental and observational data, among others. 
In addition, a novel search-operator is proposed to take advantage of such prior knowledge.
Experiments show that, using path beliefs improves the learning of the skeleton, as well as the edge directions in the network.
\end{abstract}

\section{INTRODUCTION}
One theoretical advantage of the search-and-score approach to learning Bayesian Networks \citep{cooper} versus the constraint-based approach \citep{Spirtes2000} is that the former naturally accepts priors for each network. 
Since the number of possible networks is exponential to the number of nodes, in a practical setting one has to assign priors in an implicit way. 
%
In this paper, we consider prior beliefs on the possible paths between variable pairs.
Such paths directly correspond to {\bf causal} or {\bf associative} relations. 
The joint beliefs on the paths is then employed to assign a prior on each network.

Causal knowledge naturally derives from prior experimental data while associative knowledge stems from observational data. 
For example, consider a dataset $\mathcal{D}$ measuring the average amount of exercise per week $E$, calcium in diet $C$, occurrence of osteoporosis by 60yrs $O$ and smoking $S$ in a cohort of women. 
A Bayesian Network could be induced by any appropriate learning method. 
However, if a prior {\em experimental study} showed that increasing the amount of exercise reduces the occurrence of the disease, then the knowledge belief that [$E$ causes (that is, causally affects) $O$] with probability $p$ should be incorporated during learning. 
Similarly, if a prior cohort study (observational study) has shown that smoking correlates with reduced exercising, then knowledge [$S$ and $E$ are associated] with probability $p'$ should also be included. 
The belief strengths $p$ and $p'$ depend on several factors, such as the statistical power of the study.
Notice that the fact [$E$ causes $O$] {\em does not} correspond to the presence of the edge $E \rightarrow O$ in the network: the edge implies a {\em direct}  causal relation (in some context of modeled variables) while [$E$ causes $O$] does not depend on the context.

{\em Path beliefs are inherently dependent}.
For example, if one believes with certainty that [$X$ causes $Y$] and [$Y$ causes $Z$], then one has to believe that [$X$ causes $Z$] to be consistent.
Therefore, one should consider the joint distribution of the input path beliefs, instead of the marginal distributions separately.
However, it is very unlikely that the complete joint distribution is available.
Instead, one could use the marginal distributions to infer a joint distribution. However, there are several technical difficulties to consider.
For example, assume we are given $P(X\text{ causes }Y) = 0.8$ and $P(Y\text{ causes } Z) = 0.8$ and wish to compute $P(X\text{ causes }Y, Y\text{ causes } Z)$.
On one hand, there may be several choices for the joint given the same marginal beliefs. 
In the above scenario we can infer $P(X\text{ causes }Y, Y\text{ causes } Z) \in [0.6, 1]$.
On the other hand, the beliefs maybe \textit{incoherent} \citep{Hansen2000}, that is, not extendable to a joint distribution that satisfies the probability axioms. 

We present a method that computes a joint distribution of the path beliefs such that: if the path beliefs are coherent the joint is the closest to uninformative priors; if they are incoherent the joint is chosen to be coherent and induces path probabilities that are close to the input beliefs. 
Once the joint is computed, it can be employed to efficiently compute the prior of a network. 
Furthermore, to take advantage of the prior knowledge, we introduce a novel search-operator.

In simulated proof-of-concept experiments we show that the new scoring method can indeed take advantage of prior knowledge. 
When provided with causal knowledge, it is able to better learn {\em the orientations} of the edges and the causal relations. 
Informative priors can also facilitate learning {\em the skeleton} of the network.
Finally, we show that the proposed search-operator significantly improves the quality of the learned model.

There are several other methods that make use of prior knowledge when learning a network (see \citep{Angelopoulos08} for a review). For example, using knowledge regarding the parameters of the network \citep{MitchellJMLR}, a causal total order of the variables \citep{cooper}
, the presence or absence of directed edges in the network \citep{meek} possibly with beliefs assigned to them \citep{Buntine,Castelo2000},
or a prior network, used to assign prior probabilities to each network based on the distance from this network \citep{heckerman}.
In general, it can be argued that the type of knowledge the existing methods can incorporate during learning is not in a form that can be easily acquired. 
As a result, uniform - and thus uninformative - priors are commonly used when learning Bayesian Networks from data. {\em The problem of incorporating informative priors while learning is listed in the list of open problems in a recent causality editorial} \citep{SpirtesJMLR}.

There also is prior work that specifically considers path constraints or beliefs. 
The methods in \citep{esann2011,BorboudakisICML2012} assume one {\em first learns} a Markov-Equivalence class of Bayesian Networks or Maximal Ancestral Graphs \citep{Spirtes2000} (a generalization of Bayesian Networks that admits hidden variables)  from data and {\em then}, path constraints are imposed on the graph. 
In contrast, in this work the network is learned {\em with the help of the prior knowledge}. 
In \citep{O'donnell2006} a method is presented for incorporating beliefs on paths, but relies on computationally expensive Markov Chain Monte Carlo (MCMC) simulations. 
However, neither the latter, nor any other method dealing with prior knowledge deals with the issues of dependent and possibly incoherent beliefs.

\section{BACKGROUND}
\label{sec:background}

We assume the reader's familiarity with Bayesian Networks \citep{Pearl2000, Neapolitan2003} and learning algorithms and just briefly review the basic concepts. 

Let $\mathcal{V}$ be a set of $k$ random variables $\{{V_i}\}_{i=1}^{k}$. 
A \textbf{Bayesian Network} (BN) over $\mathcal{V}$ is a pair $\mathcal{B = \langle G_{V}, P_{V}} \rangle$, where $G_{\mathcal{V}}$ is a \textbf{Directed Acyclic Graph} (DAG) representing conditional independencies between variables $\mathcal{V}$, and $P_{\mathcal{V}}$ is the joint probability distribution (j.p.d.) of $\mathcal{V}$. The graph and distribution must be connected by the equation $P_{\mathcal{V}} = \prod{P(V_i | Pa_{\mathcal{G}}(V_i))}$, where $Pa_{\mathcal{G}}(V_i)$ are the parents of $V_i$ in $\mathcal{G}$. The above equation is equivalent to what is called the \textbf{Markov Condition}. When the network is fixed in a context we drop the indexes $\mathcal{V}, \mathcal{G}$ from the equations.
The \textbf{skeleton} of a BN $\mathcal{G}$ is the undirected graph which can be constructed by ignoring the orientations of $\mathcal{G}$. 
A triple of vertices $\langle X,Y,Z \rangle$ is called a \textbf{collider} in $\mathcal{G}$, if $X \rightarrow Y \leftarrow Z$ is in $\mathcal{G}$. 
A collider $\langle X,Y,Z \rangle$ is \textbf{unshielded} if $X$ and $Z$ are not adjacent in $\mathcal{G}$. 
Two BNs are called \textbf{Markov equivalent} if: (a) they have the same skeleton, and (b) they have the same set of unshielded colliders. 
A \textbf{Partially Directed Acyclic Graph} (PDAG) (also known as essential graph) is a graph representing a set of Markov equivalent BNs. 
It has the same skeleton as all BN representatives and an edge is directed if and only if it is invariant in all BN representatives.
A {\bf directed path} from $X$ to $Y$ is denoted as $X \rightpath Y$. 
We {\bf denote as $X \confounded Y$} the case where $X$ and $Y$ share a common ancestor in $\mathcal{G}$, but neither $X$ is an ancestor of $Y$ nor the reverse.  A {\bf $d$-connecting path} (given the empty set) between $X$ and $Y$ exists if either $X \rightpath Y$, $X \leftpath Y$, or $X \confounded Y$. The absence of a $d$-connecting path between $X$ and $Y$ is {\bf denoted as $X \nopath Y$}. In the rest of the paper, we assume the Faithfulness Condition \citep{Spirtes2000} that (together with the Markov Condition) implies that {\em there is a $d$-connecting path between $X$ and $Y$, if and only if the two nodes are statistically associated (dependent)}. 

Let $\mathcal{D}$ be a complete multinomial dataset over variables $\mathcal{V}$.
The probability of a network $\mathcal{G}$ over $\mathcal{V}$ is $P(G|D) \propto P(D|G) \cdot P(G)$.
The score of a network is often obtained by taking the logarithm of $P(G | D)$, and equals $Sc(G | D) = Sc(D | G) + Sc(G)$ .
Bayesian scoring methods such as K2 \citep{cooper} and BDe, BDeu, \citep{heckerman} try to approximate the log-likelihood based on different assumptions. 
When priors are uniform, the term $Sc(G)$ can be ignored during maximization.
In our setting however, this term may become important.

\section{REPRESENTING PATH BELIEFS}

For any pair $X,Y \in \mathcal{V}$ we may have a prior belief on the possible paths connecting the two nodes in the network.
It is important that we devise cases for such paths that are {\em mutually exclusive} and {\em allow the representation of common types of causal and associative knowledge}.
This is possible as follows: we define the \textbf{path variables} $r_{i,j}$ taking values in the domain $\{\rightpath, \leftpath, \confounded, \nopath\}$ with the semantics $V_i \rightpath V_j$, $V_i \leftpath V_j$, $V_i \confounded V_j$, and $V_i \nopath V_j$ respectively.
When the specific nodes $V_i$, $V_j$ we refer to are not important we will use a single index: $r_k$.
Each variable $r_{i,j}$ has a probability distribution $\Pi_{r_{i,j}} = \langle \pi_\rightpath, \pi_\leftpath, \pi_\confounded, \pi_\nopath \rangle$ over each possible value.
The input to our method is a set of path beliefs $\mathbf{K} = \langle \mathbf{R}, \mathbf{\Pi} \rangle$, where $\mathbf{R}$ is a set of path variables and $\mathbf{\Pi}$ the set of probability distributions associated with them.
An example is shown in Table \ref{tb:marginal_coherent}(Top) expressing the belief that most likely there is a directed path from $X$ to $Y$ and from $Y$ to $Z$.

The possible paths between nodes dictate their possible \textbf{causal} and \textbf{associative} relations.
When the BN is interpreted causally, then $X \rightpath Y$ is equivalent to [$X$ causes $Y$].
In addition, as discussed in the previous section: $X \rightpath Y$ or $X \leftpath Y$ or $X \confounded Y$ is equivalent to [$X$ is associated with $Y$].
Thus, a distribution $\Pi_{r_{X,Y}} = \langle \pi_\rightpath, \pi_\leftpath, \pi_\confounded, \pi_\nopath \rangle$ corresponds to beliefs about the causal and associative relations.


In practice, it is useful to allow the user to specify prior beliefs directly on the events [$X$ does (not) cause $Y$] and [$X$ is (not) associated with $Y$] from which the distribution $\Pi_{r_{XY}}$ can be derived, than the opposite.
This is not difficult: for example given $P(X \text{ causes } Y) = \pi_\rightpath$ the mass of probability $1-\pi_\rightpath$ has to be distributed in a reasonable way to the other three values.
However, we avoid this belief representation to simplify the presentation of the method.

\begin{table*}
\centering
\caption{(a) (Top Part) Path beliefs $\mathbf{K}$ for three pairs of nodes. The beliefs are {\em incoherent}: $P(X \rightpath Y) = 0.8$ and $P(Y \rightpath Z) = 0.9$ imply that $P(X\rightpath Z) \in [0.7,1]$. (a) (Bottom Part) Induced {\em coherent} beliefs $\mathbf{K'}$ stemming from $\mathbf{K}$ by solving the problem in Eq. \ref{eq:opt1}. (b) A part of the j.p.d. $J$ computed by solving Eq. \ref{eq:opt1} with input $\mathbf{K'}$. The number of DAGs with 5 nodes for each configurations $N_C$ is also shown. 
Notice that $C_2$ and $C_3$ have both zero counts and zero probability, because they are invalid.}
\subfloat[][]{
    \small
    \label{tb:marginal_coherent}
    \begin{tabular}{| c | c | c | c | c |}
        \hline
        {\bf K} & $\pi_\rightpath$ & $\pi_\leftpath$ & $\pi_\confounded$ & $\pi_\nopath$ \\ \hline
        $r_{X,Y} (r_1)$ & 0.8 & 0.132 & 0.028 & 0.04 \\ \hline
        $r_{Y,Z} (r_2)$ & 0.9 & 0.066 & 0.014 & 0.02 \\ \hline
        $r_{X,Z} (r_3)$ & 0.6 & 0.264 & 0.056 & 0.08 \\
        \hline
        \hline
        {\bf K$'$} & $\pi_\rightpath$ & $\pi_\leftpath$ & $\pi_\confounded$ & $\pi_\nopath$ \\ \hline
        $r_{X,Y} (r_1)$ & 0.764 & 0.159 & 0.032 & 0.045 \\ \hline
        $r_{Y,Z} (r_2)$ & 0.879 & 0.082 & 0.016 & 0.023 \\ \hline
        $r_{X,Z} (r_3)$ & 0.646 & 0.231 & 0.051 & 0.073 \\
        \hline
        \end{tabular}
        }
\subfloat[][]{
\small
\label{tb:joint}
        \begin{tabular}{| l | c | c | c | c | c |}
        \hline
        & $r_{X,Y} $ & $r_{Y,Z}$ & $r_{X,Z}$ & $J_C$ & $N_C$ \\ \hline
        $C_1$ & $\rightpath$ & $\rightpath$ & $\rightpath$ & 0.6443 & 2800 \\ \hline
        $C_2$ & $\rightpath$ & $\rightpath$ & $\leftpath$ & 0 & 0 \\ \hline
        $C_3$ & $\rightpath$ & $\rightpath$ & $\confounded$ & 0 & 0 \\ \hline
        $ \dots $ & \dots & \dots & \dots & \dots & \dots \\ \hline
        $C_{49}$ & $\nopath$ & $\rightpath$ & $\rightpath$ & $4.55 \cdot 10^{-4}$ & 1045 \\ \hline
        $ \dots $ & \dots & \dots & \dots & \dots & \dots \\ \hline
        $C_{64}$ & $\nopath$ & $\nopath$ & $\nopath$ & $2.78 \cdot 10^{-5}$ & 309 \\
        \hline
        \end{tabular}
        }
\end{table*}

\section{SCORING PATH BELIEFS}

In this section, we derive a score for DAG $G$ given data $D$ and $n$ path beliefs in $\mathbf{K}$.
An important requirement for the computation of the score is knowledge of a joint distribution $J = P(r_1, \ldots, r_n | \mathbf{\Pi}) = P(\mathbf{R} | \mathbf{\Pi})$ such that its marginals correspond to the distributions in $\mathbf{\Pi}$.
We assume $J$ is already computed; the following sections describe the details of this computation.
The j.p.d. $J$ stemming from $\mathbf{K}$ in Table \ref{tb:marginal_coherent} is shown in Table \ref{tb:joint}.

We denote with $C$ (configuration) a given joint instantiation of values to path variables $\mathbf{R} = \langle r_1, \ldots r_n \rangle$, and define $J_C =  P(\mathbf{R} = C | \mathbf{\Pi})$.
It is important to notice that {\em for each graph $G$ the configuration $C$ is uniquely determined}.
For example, in the j.p.d. of Table \ref{tb:joint}, if in a graph $G$ $X \rightpath Y$, $Y \rightpath Z$ and $X \rightpath Z$ hold, then ${\bf r} = C_1$.
Thus, it makes sense to denote with $C_G$ the joint instantiation of variables $\mathbf{R}$ in graph $G$.

%

Let $G$ be a DAG and $D$ a dataset over the same variables. We now compute the probability $P(G | D, J)$:
\begin{equation*}
P(G | D, J) =\frac{P(D | G, J) \cdot P(G | J)}{P(D | J)}
 =\frac{P(D | G) \cdot P(G | J)}{P(D | J)}
\end{equation*}
The second equation stems from the fact that given the graph $G$ the data $D$ are independent of $J$ ($J$ does not provide any additional information about the data once the graph is known). The factor $P(D | J)$ is a normalizing constant that does not need be computed when we maximize the above equation over different graphs. In Section \ref{sec:background} we mention several approximations for computing the factor $P(D | G)$.
 We now focus on the prior $P(G | J)$:
\begin{align*}
P(G | J) = & P(G, C_G| J) = P(G | J, C_G) \cdot P(C_G | J) =  \\
& P(G | C_G) \cdot P(C_G | J) = P(G | C_G) \cdot J_{C_G}
\end{align*}

The first equation holds because $C_G$ is a function of $G$.
The factor $P(G | C_G)$ is our prior on a graph $G$ given that a specific configuration holds. Given no other preference or knowledge {\em we assign the same prior to all graphs with the same configuration}. 
Let $N_C$ be the number of DAGs over nodes $\mathcal{V}$ sharing the same configuration $C$.
Then $P(G | C_G) = 1/N_{C_G}$ and so:
\begin{equation}
\label{eq:prior}
{
P(G | J) = \frac{J_{C_G}}{N_{C_G}}
}
\quad\text{and}\quad
{
Sc(G | J) = \log\left(\frac{J_{C_G}}{N_{C_G}}\right)
}
\end{equation}
The overall score of a graph is then defined as:
\begin{equation}
\label{eq:priorscore}
{
Sc(G | D, J) = Sc(D | G) + Sc(G | J)
}
\end{equation}
The score $Sc(G | D, J)$ has two desirable properties:
\begin{enumerate}
\item{{\bf Markov-Equivalent graphs that satisfy the same path-beliefs obtain the same score.} The last term in the equation above is the same for graphs sharing the same configuration. The first term is the same for Markov-equivalent graphs provided one employs an appropriate scoring function, such as the BDe score \citep{heckerman}.}
\item{\label{la:condition2} {\bf For uninformative prior beliefs, all graphs are equiprobable a priori}, that is, $P(G | J) = 1 / N$, where $N$ is the number of graphs over nodes $\mathcal{V}$. With uninformative beliefs we expect to encounter a given configuration with probability equal to the proportion of the graphs satisfying the configuration, i.e,. $J_C = \frac{N_C}{N}$. In that case, $P(G | J) = \frac{N_C}{N} \cdot \frac{1}{N_C} = \frac{1}{N}$ and we end up with uniform priors as we would expect.}
\end{enumerate}

While Eq. \ref{eq:prior} follows the above two properties, we point out to the fact that the factor ${1}/{N_{C_G}}$ may seem to provide counter-intuitive results at a first glance.
The reason is that, everything else being equal, higher priors will tend to be assigned to graphs in ``small" configurations, that is, consistent with only a few graphs.
If this is not desirable then one can drop the $1/N_C$ factor. 
However, if this score is used in place of Eq. \ref{eq:prior} then Property \ref{la:condition2} above is not satisfied any more.

\section{COMPUTING THE NUMBER OF DAGS $N_C$}

The number $N$ of DAGs over nodes $\mathcal{V}$ has been solved in closed-form \citep{Robinson}.
However, to the best of our knowledge, there is no closed-form for the number $N_C$ of DAGs that satisfy certain path-constraints. 
When the number of nodes is small (up to 5-6) one can enumerate all DAGs and compute each $N_C$ by counting.
The number of possible DAGs however, grows exponentially to the number of nodes and complete enumeration is not an option.
In this case, we estimate these counts by sampling a number $S$ of DAGs uniformly at random.
Specifically, we implemented the recent method in \citep{Kuipers2012} that, unlike prior work \citep{Melancon00}, avoids the use of expensive MCMC methods. 
$\hat{N}_C$ can be estimated as $N \cdot S_C/S$, where $S_C$ is the number of sampled DAGs that conform to configuration $C$.

When the number of configurations $c$ is large or $N_C / N$ is small, one may never sample any graph consistent with $C$, resulting in zero estimates.
This may happen even for small sets of path variables, as $c$ grows exponentially with the number of path variables $n$.
To avoid zero estimates, one can apply the Laplace correction: $\hat{N}_C = \frac{S_C + l}{S + cl}N$, where $l$ is an arbitrary parameter.
We suggest $l$ to be close to zero.
Later on we will refer to this method as $\mathrm{FULL}_l$.

In order to get a good estimate of $N_C$ using $\mathrm{FULL}_l$, one may have to sample a huge number of DAGs.
To improve upon this we developed another method to approximate $N_C$.
This method is based on the observation that, often, certain subsets of path variables are ``almost independent".
We exploit this to factorize the uninformative prior distribution $U$ of each configuration, denoted with $U_C$ for configuration $C$. $N_C$ can then be computed as $U_C \cdot N$.
\subsection{FACTORING THE UNINFORMATIVE PRIOR DISTRIBUTION $U$}

To give an intuitive understanding of the main idea, consider the following scenario:
we are given two path variables, $r_{X,Y}$ and $r_{W,Z}$.
Notice that they do not have any nodes in common.
Assume that we fix the value of $r_{W,Z}$.
Depending on that value, some values of $r_{X,Y}$ will become more or less likely.
For example, if $W \nopath Z$ holds, the values $X \rightpath Y$, $X \leftpath Y$ and $X \confounded Y$ become less likely since $W \nopath Z$ restricts the graph to contain fewer edges, effectively reducing the possibility to form paths between $X$ and $Y$. On the other hand, $X \nopath Y$ becomes more likely.
%
 To put it formally, $U_{X \rightpath Y | W \nopath Z} < U_{X \rightpath Y}$, $U_{X \leftpath Y | W \nopath Z} < U_{X \leftpath Y}$, $U_{X \confounded Y | W \nopath Z} < U_{X \confounded Y}$ and $U_{X \nopath Y | W \nopath Z} > U_{X \nopath Y}$.
However, we claim that \textit{if the number of nodes $\mathcal{V}$ is sufficiently large}, the difference is negligible, or formally that $U_{r_{X,Y} | r_{W,Z}} \simeq U_{r_{X,Y}}$, that is, they are ``almost independent".
We illustrate this with a simple example.
Assume that $\mathcal{V} = \{X,Y,W,Z\}$.
In this case it is clear that any value of $r_{W,Z}$ heavily constrains the graph, since it only contains 4 nodes.
If however we keep adding nodes to $\mathcal{V}$, more and more possibilities are created to satisfy any value of $r_{X,Y}$.

Next we show an example with dependent path variables.
We are given the path variables $r_{X,Y}$ and $r_{Y,Z}$.
Notice that $Y$ appears in both path variables.
Now consider the configurations $C_1 = \{X \rightpath Y, Y \rightpath Z\}$ and $C_2 = \{X \rightpath Y, Y \leftpath Z\}$.
Note that the prior probability of a directed path between any two nodes is equal for any pair of nodes.
Assuming $U$ can be factorized, $U_{Y \rightpath Z | X \rightpath Y} = U_{Y \rightpath Z}$, and $U_{Y \leftpath Z | X \rightpath Y} = U_{Y \leftpath Z}$ hold.
Because $U_{Y \rightpath Z} = U_{Y \leftpath Z}$ holds, $U_{Y \rightpath Z | X \rightpath Y} = U_{Y \leftpath Z | X \rightpath Y}$ follows.
However, given $X \rightpath Y$, $Y \rightpath Z$ becomes less likely since there are no DAGs with $Z \rightpath X$ (acyclicity), which is not the case for $Y \leftpath Z$.
For example, for $\mathcal{V} = \{X,Y,Z\}$ there are only 2 DAGs with configuration $C_1$, but 4 DAGs with configuration $C_2$.
Thus $U$ cannot be factorized in this case.


Those scenarios only give a rough and intuitive understanding of the basic idea.
In the next subsection we will provide experimental results to support our claims.
Before doing so, we will generalize the basic ideas to any set of path beliefs.


%

%

\begin{definition}
Let $\mathbf{R}$ be a set of path variables.
We denote with $V_R$ the set of all nodes appearing in any variable in $\mathbf{R}$.
The \textbf{constraint graph} $G_R = (V_R, E_R)$ of $\mathbf{R}$ is an undirected graph, where $E_R = \{X - Y\}_{r_{X,Y} \in \mathbf{R}}$.
\end{definition}

\begin{definition}
Let $\mathbf{R}$ be a set of path variables and $\mathbf{P}$ a partition of $\mathbf{R}$.
Let $\mathcal{V}_{R_i}$ denote the set of all nodes appearing in any variable in the $i$-th part of $\mathbf{P}$, $\mathbf{P}_i$.
$\mathbf{P}$ is called an \textbf{independent partition} if $\forall \mathbf{P}_i, \mathbf{P}_j \in \mathbf{P}, i \neq j, \mathcal{V}_i \bigcap \mathcal{V}_j = \emptyset$.
\end{definition}

Since the parts of an independent partition do not have any nodes in common, the configuration of a part does not \textit{directly} influence any other part; they do however have an \textit{indirect} influence through other nodes of the graph which, as we will see, is negligible.
On the other hand, path variables of the same part do directly affect each other (see dependent case above).

The independent partition of a set of path variables $\mathbf{R}$ can be computed as follows: (a) construct the constraint graph $G_R$ of $\mathbf{R}$ and, (b) find the connected components of $G_R$. It is easy to see that the connected components of $G_R$ are an independent partition of $\mathbf{R}$.

%


It remains to show how to compute $U$ for given set of path variables $\mathbf{R}$ and set of nodes $\mathcal{V}$.
First we sample $S$ DAGs over $\mathcal{V}$ uniformly at random.
Then we find an independent partition $\mathbf{P}$ of $\mathbf{R}$.
$U_C$ is factorized as $U_C = \prod_i U_{C_i}$, where $C_i$ and $U_{C_i}$ denote the configuration and the prior distribution of the $i$-th part $\mathbf{P}_i$ of $\mathbf{P}$ respectively.
$U_{C_i}$ is estimated as $S_{C_i}/{S}$, where $S_{C_i}$ is the number of sampled DAGs that conform to configuration $C_i$ in $\mathbf{P}_i$.
Finally, $\hat{N}_C = N \cdot U_C$.
Again, we recommend a Laplace correction.
Then, $\hat{N}_C = N \cdot \frac{S \cdot U_C + l}{S + l \cdot c}$.
We will refer to this method as $\mathrm{FACT}_l$.

\subsection{EXPERIMENTAL VALIDATION}

\begin{figure*}[ht]
\centerline{
\subfloat[][]{
\label{fig:uninformative_nodes}
\includegraphics[width=0.49\columnwidth]{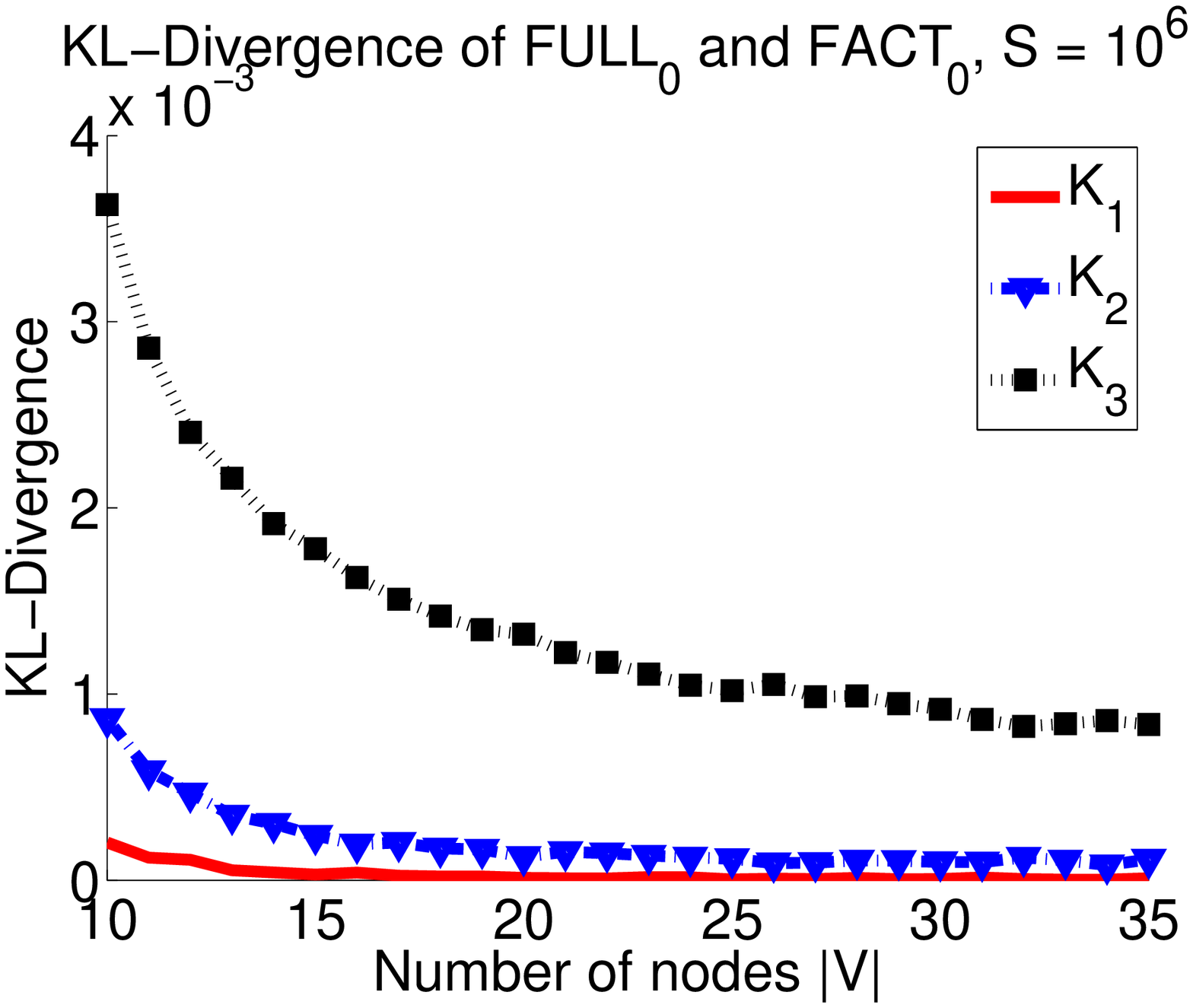}
}
\subfloat[][]{
\label{fig:uninformative_samples_4}
\includegraphics[width=0.49\columnwidth]{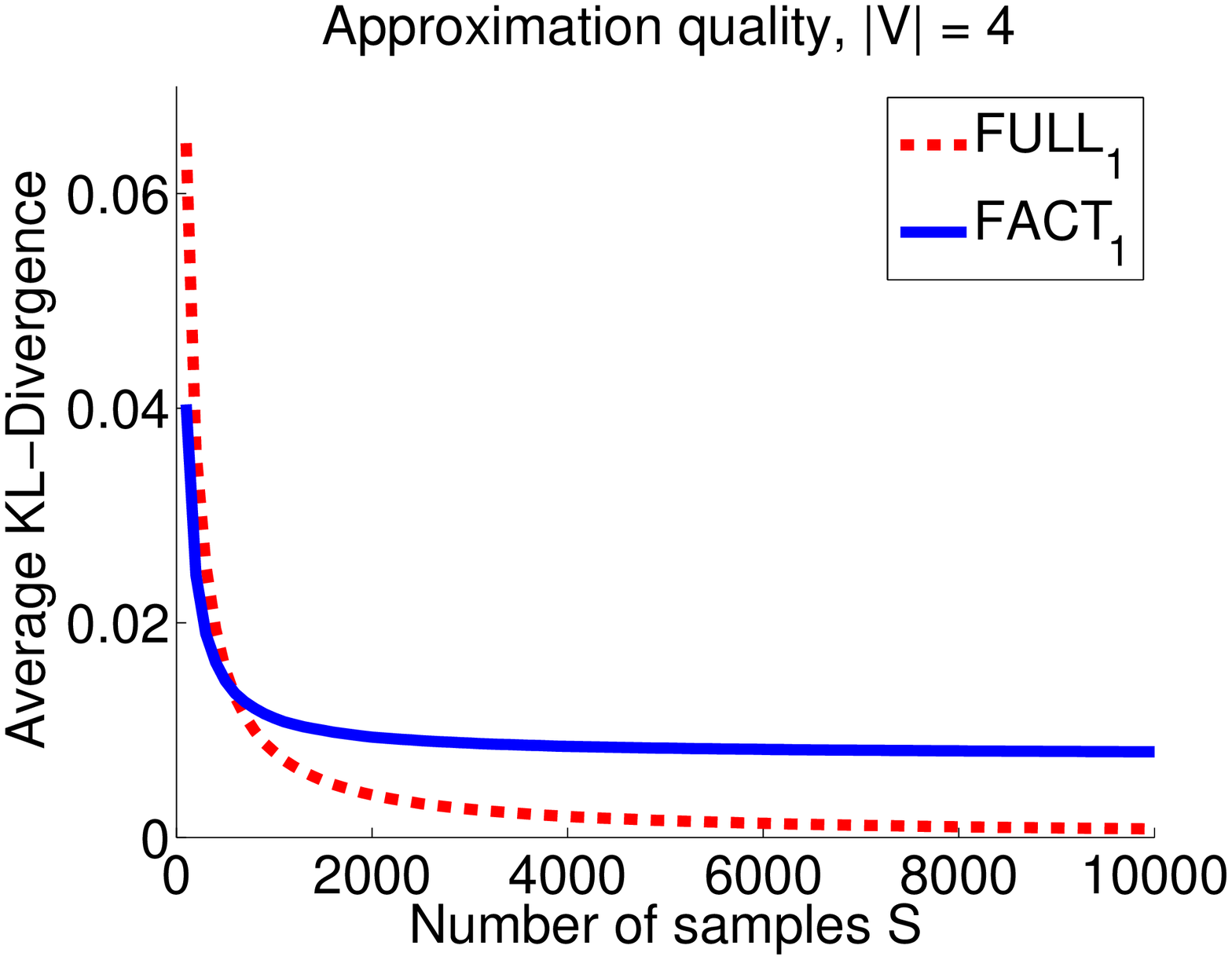}
}
\subfloat[][]{
\label{fig:uninformative_samples_5}
\includegraphics[width=0.49\columnwidth]{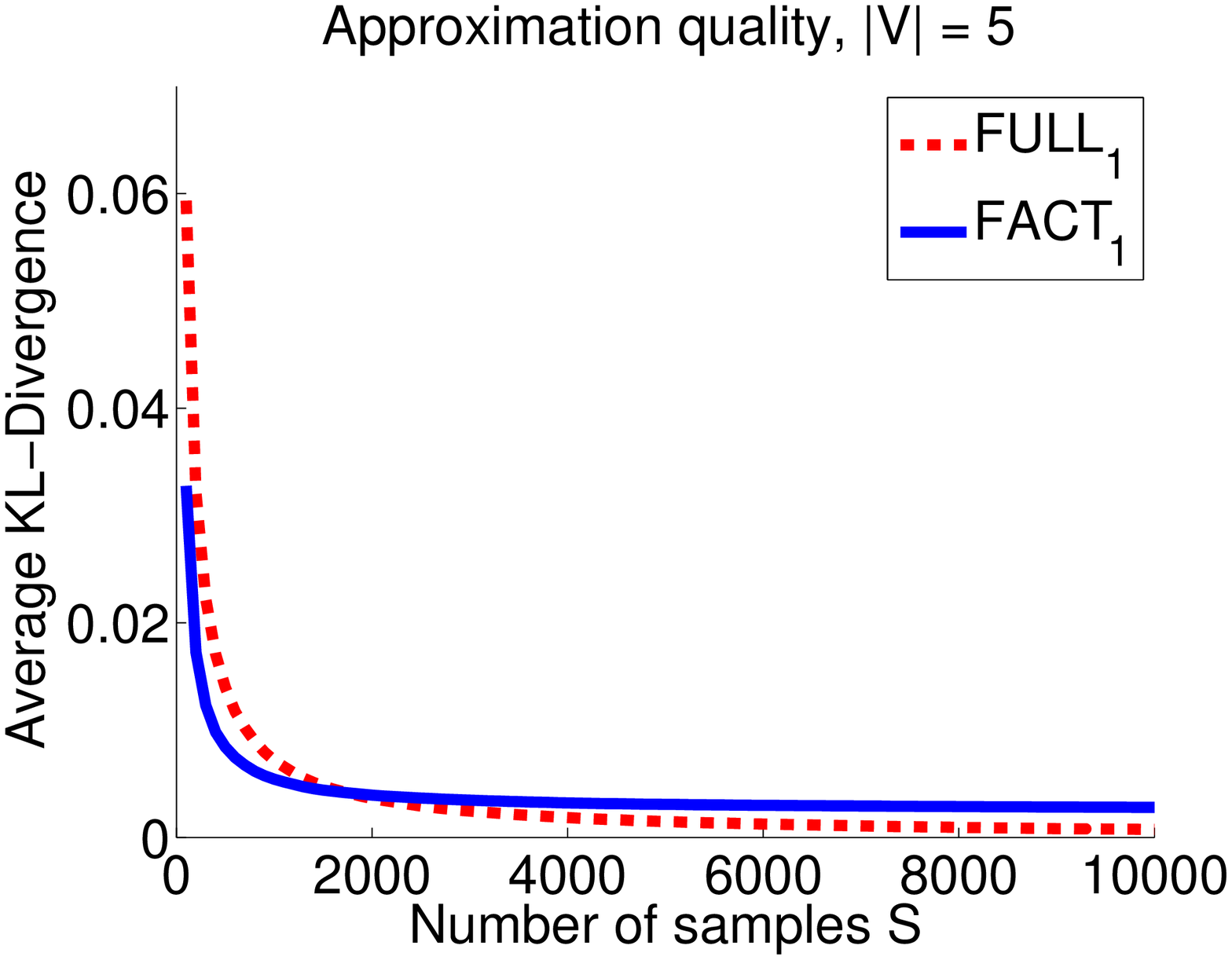}
}
\subfloat[][]{
\label{fig:uninformative_samples_6}
\includegraphics[width=0.49\columnwidth]{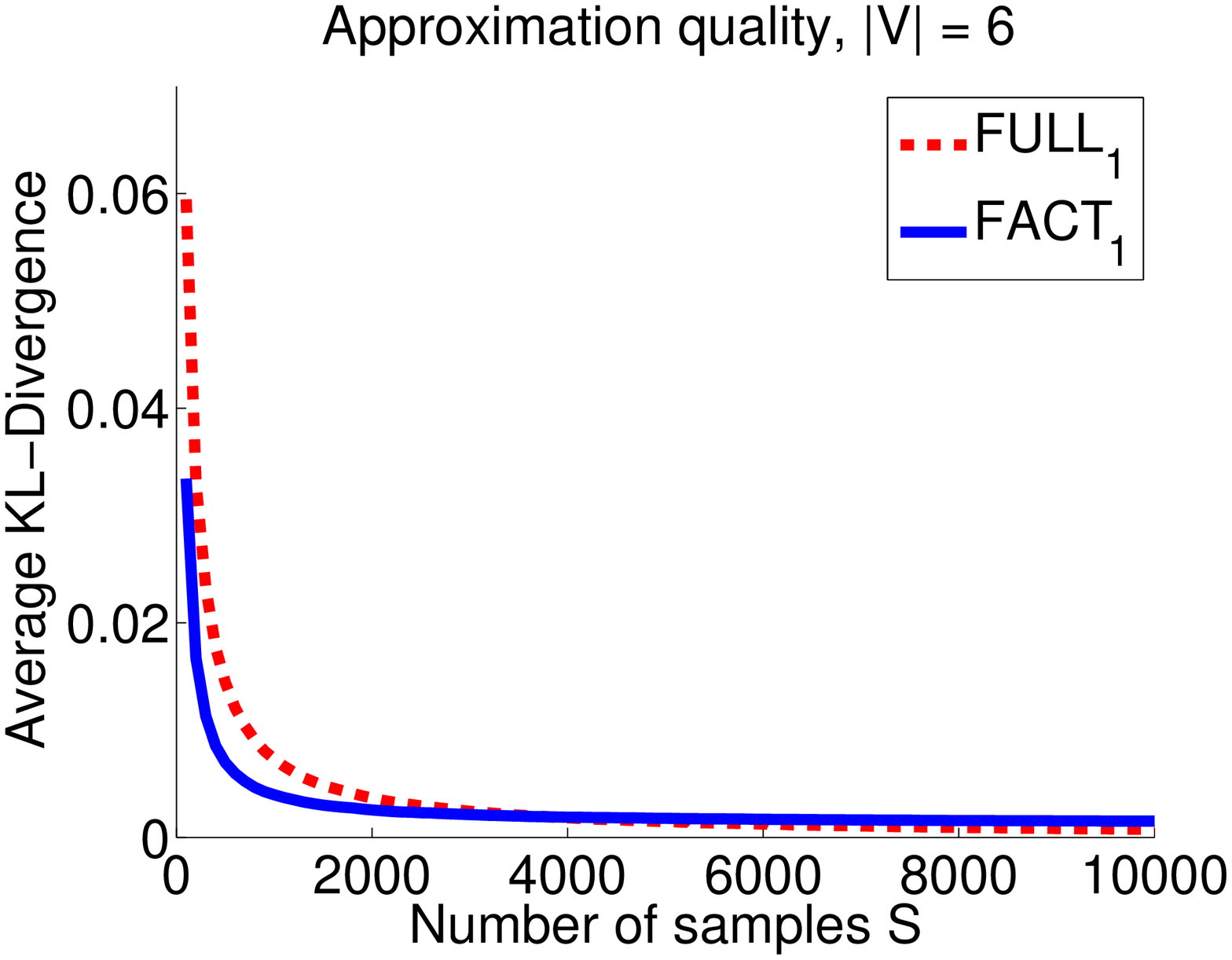}
}
}
\caption{(a) KL-divergence between $\mathrm{FULL}_0$ and $\mathrm{FACT}_0$ with $S = 10^6$ for different sets of path variables. The distance between $\mathrm{FACT}_0$ and the true distribution, approximated by $\mathrm{FACT}_0$, decreases as the number of nodes increases. (b,c,d) KL-divergence between the true distribution and the approximation methods, as the number of samples S increases. For small S $\mathrm{FACT}_1$ provides a better approximation of the true distribution than $\mathrm{FULL}_1$.}
\label{fig:uninformative}
\end{figure*}%

The first experiment is to determine how $\mathrm{FACT}_l$ approximates $\mathrm{FULL}_l$, as the number of nodes $|\mathcal{V}|$ increases.
We denote with $U_{\mathrm{FACT}_l}$ and $U_{\mathrm{FULL}_l}$ the estimation of  $U$ by the methods $\mathrm{FACT}_l$ and $\mathrm{FULL}_l$.

\textbf{Setup}: The number of nodes is varied between $10$ and $35$, with a step-size of $1$.
We used three sets of path variables: $R_1 = \{r_{1,2}, r_{3,4}\}$, $R_2 = \{ r_{1,2}, r_{2,3}, r_{4,5}, r_{5,6}\}$, and $R_3 = \{r_{1,2}, r_{2,3}, r_{3,4}, r_{5,6},r_{6,7},r_{7,8}\}$.
The number of independent partitions is 2 and each paritition consists of 1,2 and 3 path beliefs for $R_1$, $R_2$ and $R_3$ respectively.
The number of valid configurations $c$ is $16$, $256$ and $1681$ for $R_1$, $R_2$ and $R_3$ respectively.
The number of sampled DAGs $S$ was set to $10^6$, sufficiently large for $\mathrm{FULL}_l$ to approximate $U$ well.
The Laplace correction parameter $l$ was set to 0, since no correction is necessary in this case.
We used the KL-divergence to measure the distance between two probability distributions, with $U_{\mathrm{FULL}_l}$ representing the true distribution.
%

\textbf{Results}:
The results are shown in Figure \ref{fig:uninformative_nodes}.
As claimed, for a fixed set of path beliefs, $U_{\mathrm{FACT}_l}$ approaches $U_{\mathrm{FULL}_l}$ (which should be close to $U$ in this experiment, as $S$ is large relative to $c$) as the number of nodes increases.
Similar results are expected with more and larger independent partitions.

In the second experiment we show that if $S$ is relatively small compared to $c$, \textit{$\mathrm{FACT}_l$ provides a better approximation of $U$ than $\mathrm{FULL}_l$}.
This is important because sampling a large number of DAGs costs time and memory, essentially setting an upper limit to $S$ which, if $c$ is relatively large, will result in a poor approximation of $U$ by $\mathrm{FULL}_l$.
To show this, one has to know the exact distribution $U$.
However, as this is computationally infeasible for large numbers of nodes, we ran the experiment only for small $|V|$.

\textbf{Setup}: The number of nodes is $|\mathcal{V}| = \{4,5,6\}$, and the number of DAGs is $543$, $29281$ and $3781503$ respectively.
Because $|\mathcal{V}|$ is small, we used only two path variables
$\mathbf{R} = \{ r_{1,2}, r_{3,4}\}$.
For each $\mathcal{V}$ we sampled between $100$ and $10000$ DAGs, with a step-size of $100$.
This was done to simulate the case where no access to the complete set of DAGs is given.
The Laplace correction constant $l$ was set to $1$.
For each $|\mathcal{V}|$ and $S$ we measured the KL-divergence between $U_{\mathrm{FULL}_l}$ and $U$, as well as between $U_{\mathrm{FACT}_l}$ and $U$.
The experiment was repeated $1000$ times and averages are reported.

\textbf{Results}:
The results are shown in Figures \ref{fig:uninformative_samples_4} to \ref{fig:uninformative_samples_6}.
When $S$ is small, $\mathrm{FACT}_l$ provides a better approximation of $U$ than $\mathrm{FULL}_l$.
The reason this works is that, if $\mathbf{R}$ is partitioned into multiple sets, each containing a relatively small number of path variables, their distributions are easier to approximate.


\section{COMPUTING THE J.P.D. $J$}
In this section, we show how to compute the joint probability distribution $J$. 
We denote with $\pi_{k,j}$ the probability that $r_k$ takes value $j \in \{ \rightpath, \leftpath, \confounded, \nopath \}$: $\pi_{k,j} = P(r_k = j)$.
The {\em unknown quantities} are $J_C$ for each configuration $C$ in $J$.
Let $\mathcal{C}_{k,j} = \{C, \text{ s.t. } r_k = j\}$, that is, the set of configurations where variable $r_k$ obtains value $j$.
For each $k$ and $j$ we obtain the following constraints:

\vspace{-0.4cm}
\begin{equation}
\label{eq:marginals}
\pi_{k,j} = \sum_{C \in \mathcal{C}_{k,j}} J_C
\end{equation}

In other words, the marginals of the j.p.d. should equal our input path beliefs.
Recall that {\em path beliefs are not independent in general.}
Thus, it is important to consider the following constraints, stemming from the path semantics of the variables {\bf R}:

\vspace{-0.4cm}
\begin{equation}
\label{eq:strzeros}
J_{C} = 0, \text{ when } C \text{ is invalid}
\end{equation}

A configuration is invalid if it cannot be satisfied by any DAG over $\mathcal{V}$, for example, it contains directed cycles.
The algorithm to detect invalid configurations is discussed in Section \ref{sec:invalid}.
To complete the problem specification we impose that:

\vspace{-0.2cm}
\begin{equation}
\label{eq:probconstraints}
\sum_C J_C = 1 \quad \text{ and } \quad J_C \geq 0
\end{equation}
\vspace{-0.2cm}

If constraints in Eqs. \ref{eq:marginals}, \ref{eq:strzeros}, \ref{eq:probconstraints} can be satisfied then a j.p.d. adhering to the probability axioms can be found such that the prior marginal beliefs hold. In this case, by definition, ${\bf K}$ is {\em coherent}, otherwise it is {\em incoherent}.

\subsection{THE CASE OF COHERENT BELIEFS}

The system of equations contains $4n$ constraints from Eq. \ref{eq:marginals}, 1 constraint from Eq. \ref{eq:probconstraints} and $c = O(4^n)$ unknowns.
For most typical problems, $4n + 1 \ll c$ and so the system may have infinite solutions.
We argue that one should choose a solution j.p.d. $J$ as close to the uninformative one as possible.
Any other distribution may introduce bias towards certain configurations, even if the prior knowledge does not suggest preference over those configurations.
In other words, if the uninformative j.p.d. $U$ is a coherent extension of the path beliefs, there is no reason to prefer any other solution over it.
A natural, information-theoretic approach is to select a j.p.d. $J$ that minimizes the KL-divergence from $U$.
The problem is formulated as:

\vspace{-0.6cm}
\begin{equation}
\label{eq:opt1}
{
\min_{\bf J} D_{KL}(J \parallel U) = \sum_{k=1}^{c} J_k \cdot \ln{\frac{J_k}{U_k}}, \text{ s.t. Eqs. \ref{eq:marginals}, \ref{eq:probconstraints}}
}
\end{equation}

This optimization problem can be solved accurately and efficiently with the Iterative Scaling procedure \citep{Darroch1972, Csiszar1975}, a generalization of the Iterative Proportional Fitting Procedure (IPFP) \citep{DemingStephen1940}.

\subsection{DEALING WITH INCOHERENT BELIEFS}

In the case of incoherent beliefs there is no j.p.d. that can equal the marginal input beliefs.
Instead of requesting coherent beliefs or ignoring the incoherency, we seek for joints with marginals as close as possible to the user's input beliefs.
To solve this problem, we implemented the method proposed in \citep{Vomlel2003}, called GEMA.
GEMA is an extension of IPFP which converges even with incoherent beliefs.
In order to solve the problem it allows the marginals to change by a small amount, which is measured with the so-called $I$-aggregate criteria.
Although GEMA tends to minimize this criteria, no guarantee about its convergence to a global or local minima is provided.
We conducted some anecdotal experiments and GEMA seems to produce reasonable results.

\begin{figure*}[t]
\centerline{
\subfloat[][]{
\includegraphics[width=0.8\columnwidth]{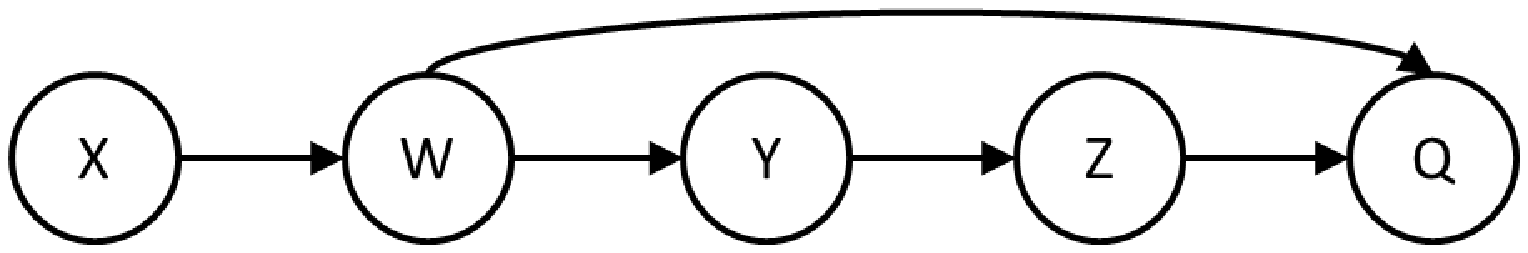}
}
\hspace{0.2\columnwidth}
\subfloat[][]{
\includegraphics[width=0.8\columnwidth]{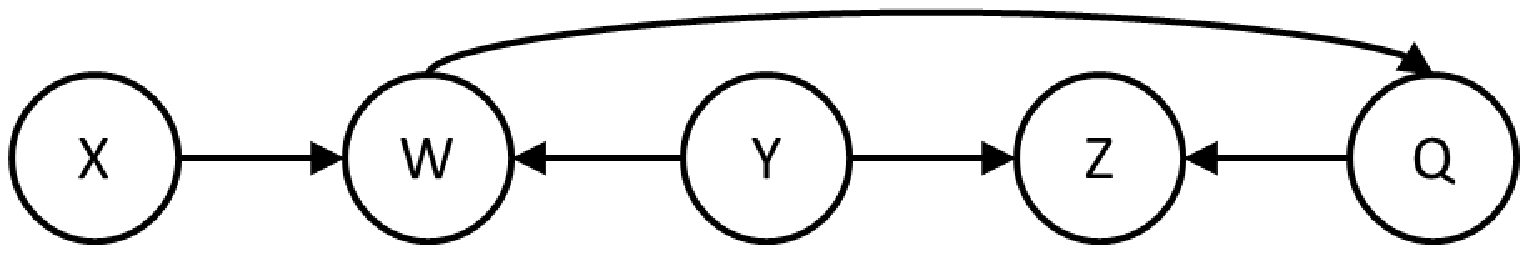}
}
}
\caption{We assume the path beliefs $\mathbf{K}$ in Table \ref{tb:marginal_coherent} and the corresponding $J$ in Table \ref{tb:joint}. (a)  The configuration $C_1 = \{X \rightpath Y, Y \rightpath Z, X \rightpath Z\}$ holds in the graph. For $p_1 = 0.6443$ we obtain the score $Sc(G_1 | J) = \log(0.6443) - \log(2800) =  -8.3769$. (b) The configuration $C_{49} = \{X \nopath Y, Y \rightpath Z, X \rightpath Z\}$ holds in the graph. For $p_{49} = 4.55 \cdot 10^{-4}$ we obtain the score $Sc(G_2 | J) = \log(4.55 \cdot 10^{-4}) - \log(1045) =  -14.6471$. As expected, $G_1$ has a higher prior than $G_2$ since $X\rightpath Y$ is given a higher probability than $X \nopath Y$ in Table \ref{tb:marginal_coherent}. }
\label{fig:example}
\vspace{-0.2cm}
\end{figure*}

Table \ref{tb:joint} contains the j.p.d. $J$ stemming from {\bf K} of Table \ref{tb:marginal_coherent}(Top) computed by GEMA.
For comparison with the input beliefs {\bf K}, Table \ref{tb:marginal_coherent}(Bottom) contains the marginal beliefs {\bf K}$'$ implied by GEMA.
The values in Table \ref{tb:marginal_coherent}(Top) and Table \ref{tb:marginal_coherent}(Bottom) are close, with the latter one representing coherent beliefs.
Figure \ref{fig:example} shows two DAGs with different configurations obtaining different prior scores.

\subsection{FACTORING THE J.P.D. $J$}

The cost of solving Eq. \ref{eq:opt1} is dominated by the number of variables $c$, which can be as high as $4^n$.
In practice, the optimization problem can not be solved efficiently (or at all, due to memory limitations) with more than 10-12 path beliefs.
It is obvious that, even in the best case, one would need at least $\Omega(c)$ time and memory, if the output of the procedure is the full j.p.d. over $c$.

One natural way to improve upon this is to factorize $J$.
Unfortunately, in general, it seems that it is not possible without loss of information.
However, as stated in \citep{Vomlel2003}, if the uninformative joint distribution $U$ factorizes with respect to some sets of variables, then the result of IPFP also factorizes with respect to the same sets of variables, that is, if $\exists \{R_i\}_{i = 1}^{k}, R_i \subseteq \mathbf{R}$ s.t. $U = \prod_{R_i} U_{R_i}$ then $J = \prod_{R_i} J_{R_i}$.
Thus, if we use $\mathrm{FACT}_l$ instead of $\mathrm{FULL}_l$ to compute $U$, we can usually compute $J$ significantly faster and for larger sets of path beliefs, that is, instead of a total limit of 10-12 path beliefs, each part of the independent partition used in $\mathrm{FACT}_l$ has a limit of 10-12 path beliefs.

\subsection{ADJUSTING MISLEADING PRIORS}\label{sec:error}
In practice, it may be the case that some priors are misleading, that is, the correct value of a path variable $r$ has a lower probability than any other value of $r$. 
It is not always possible to detect those cases; however, it is possible to do so when the path beliefs are dependent, and the majority of them gives preference to the correct relation. 
%
We illustrate this with a simple example.
Assume that the correct relation between two variables $X$ and $Y$ is $X \rightpath Y$, and that an expert suggests that $P(X \rightpath Y) = 0.1$.
Now assume that we have path beliefs that $P(X \rightpath V) = P(V \rightpath Y) = 0.9$. 
They are incoherent: by the probability axioms, $P(X \rightpath Y) \geq 0.8$ follows.
Our method will implicitly consider this and will increase $P(X \rightpath Y)$ while reducing $P(X \rightpath V)$ and $P(V \rightpath Y)$.
The effect will be even higher if more path beliefs suggest that $P(X \rightpath Y)$ is high.
For example, if $P(X \rightpath Y) = 0.1$ and we have 4 such pairs of path beliefs $P(X \rightpath V_i) = P(V_i \rightpath Y) = 0.9$, our method will assign $P(X \rightpath Y) = 0.632$ and $P(X \rightpath V_i) = P(V_i \rightpath Y) = 0.814$  $\forall i$.
We see that, although $P(X \rightpath Y)$ was low initially, it was given a high probability by our method because the other beliefs supported $X \rightpath Y$.
Thus, \emph{considering dependent beliefs and dealing with incoherence may identify and adjust 
 misleading beliefs}.

\subsection{INVALID CONFIGURATIONS}\label{sec:invalid}
%

Let $C$ be a configuration of path variables $\mathbf{R}$.
$C$ is \textbf{invalid} if $\exists r_{X,Y} \in \mathbf{R}$, s.t.:
(a) $r_{X,Y} = ``\rightpath"$ and $r_{X,Y} = ``\leftpath"$ is implied by $C$ (acyclicity), or (b) $r_{X,Y} = ``\confounded"$ and $r_{X,Y} \in \{\rightpath, \leftpath\}$ is implied by $C$ (definition of $``\confounded"$), or (c) $r_{X,Y} = ``\nopath"$ and $r_{X,Y} \in \{\rightpath, \leftpath, \confounded\}$ is implied by $C$ (definition of $``\nopath"$).

These conditions are sufficient to identify invalid configurations, but not necessary.
The simplest example is a dataset with two variables $X$ and $Y$: the configuration $r_{X,Y} = ``\confounded"$ is invalid as there is no other variable to serve as a common ancestor.
Yet, the above cases will not identify it as such.
However, when the number of variables in the data is large relative to the number of path variables (specifically if $|\mathcal{V}| \geq |V_R| + n$ holds)\footnote{There are cases where a smaller number of variables is sufficient, but we did not further investigate it.}, these conditions are also necessary.
\textit{From now on we assume that the number of nodes in $\mathcal{V}$ is sufficiently large.}

\section{SEARCH AND OPERATORS}

In this paper we will use the \textbf{Greedy Search} method, searching in the space of DAGs.
The method starts from a given initial DAG $G_0$ (usually chosen to be the empty DAG) and performs a hill-climbing search, considering all DAGs resulting by a \textbf{edge-insertion}, \textbf{edge-removal} or \textbf{edge-reversal} operation.

\subsection{EXTENDING GREEDY SEARCH}

Greedy Search can be trivially extended to additionally consider the prior score $Sc(G|J)$ of a DAG $G$.
To do this, it first has to determine the configuration $C_G$ of $G$, which can be computed in time $O(|V| \cdot n)$ given the transitive closure of $G$ (stored as an adjacency matrix).
The transitive closure of a DAG can be computed in time $O(|V|^2 + |V| \cdot |E|)$; run a DFS for each node and keep track of all visited nodes.
There are faster and more complex algorithms \citep{Simon1988}, but the trivial method is usually faster for smaller graphs (we used the trivial method in our implementation).

A problem is that, at each step of the search, the transitive closure has to be computed for all DAGs resulting by one of the search operators, whose number is $\Theta(|V|^2)$.
The total cost is then $O(|V|^4 + |V|^3 \cdot |E| + |V|^3 \cdot n)$, which is a significant computational overhead.
A straight-forward optimization is to dynamically update the closure after each edge insertion or removal.
Various methods exist \citep{Demetrescu2008} trading off the time it takes to update the closure and querying for reachability.
Assuming unit query time, a $O(|V|^2)$ update time is optimal \citep{Demetrescu2008}.
Using this method, the time-complexity can be reduced to $O(|V|^4 + |V|^3 \cdot n)$.

\subsection{SWAP-EQUIVALENT OPERATOR}

To take advantage of the extra information provided by the path beliefs, one may have to use additional search-operators.
That is because the standard operators make only small local improvements, without considering the global information provided by the path beliefs.
Thus, an operator is desirable which is able to simultaneously make multiple adjustments in order to also change the configuration of the path variables.


We propose the \textbf{swap-equivalent-operator}.
The idea is simple: at each step, after the application of a standard operator, we allow the algorithm to swap to a Markov equivalent DAG with the highest path belief score $Sc(G|J)$ increase.
If the data score $Sc(G|D)$ has the score-equivalence property (e.g. BDe), the resulting DAG has the same data score but may have a higher prior score.
This DAG can be computed with a simple modification of an algorithm presented in \citep{BorboudakisICML2012}.
Due to space limitations, the algorithm will not be described here.

\section{EXPERIMENTAL RESULTS}

\begin{figure*}
\subfloat[][]{\label{fig:exactNcollider}
\includegraphics[width=0.49\columnwidth]{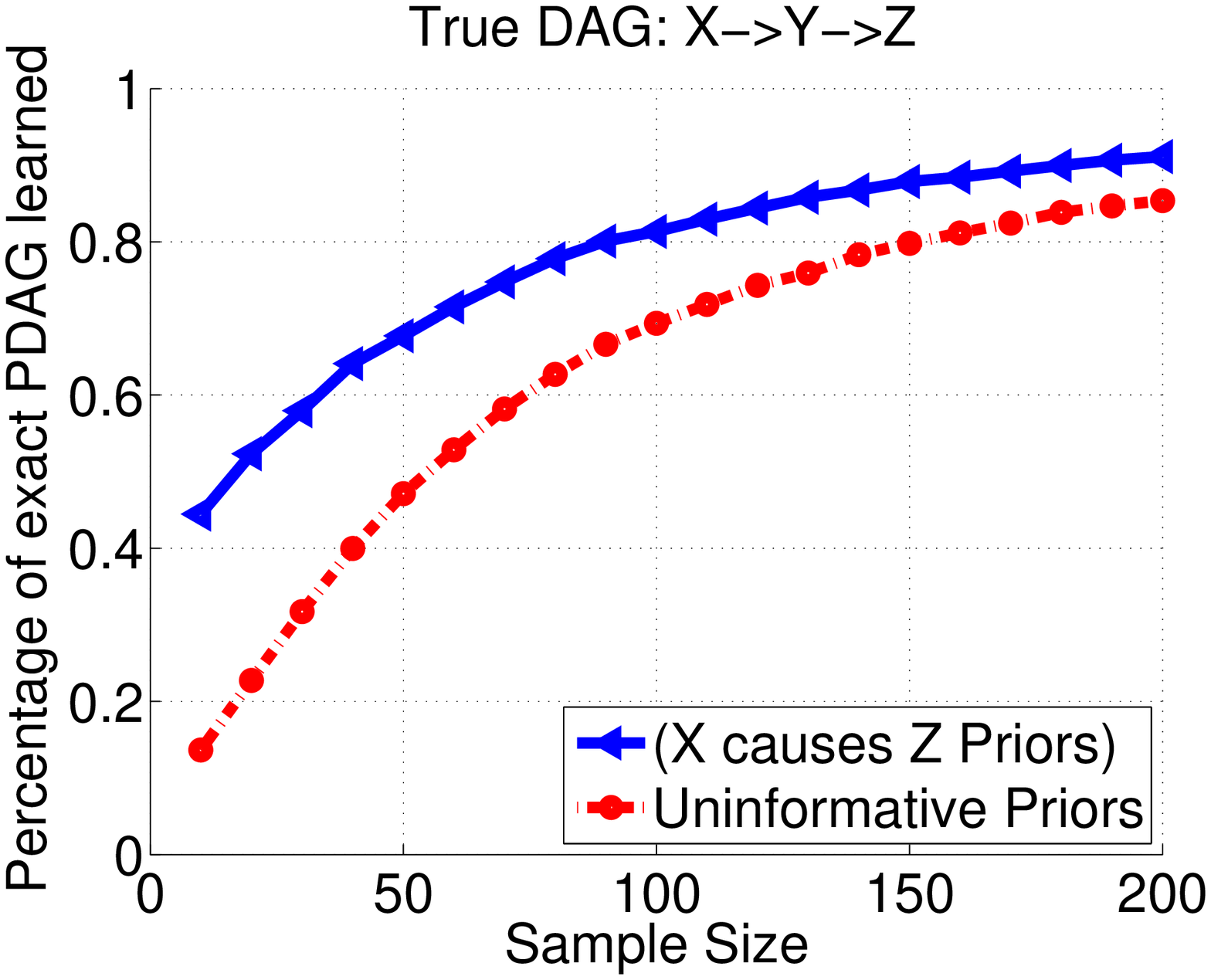}
}
\subfloat[][]{\label{fig:exactcollider}
\includegraphics[width=0.49\columnwidth]{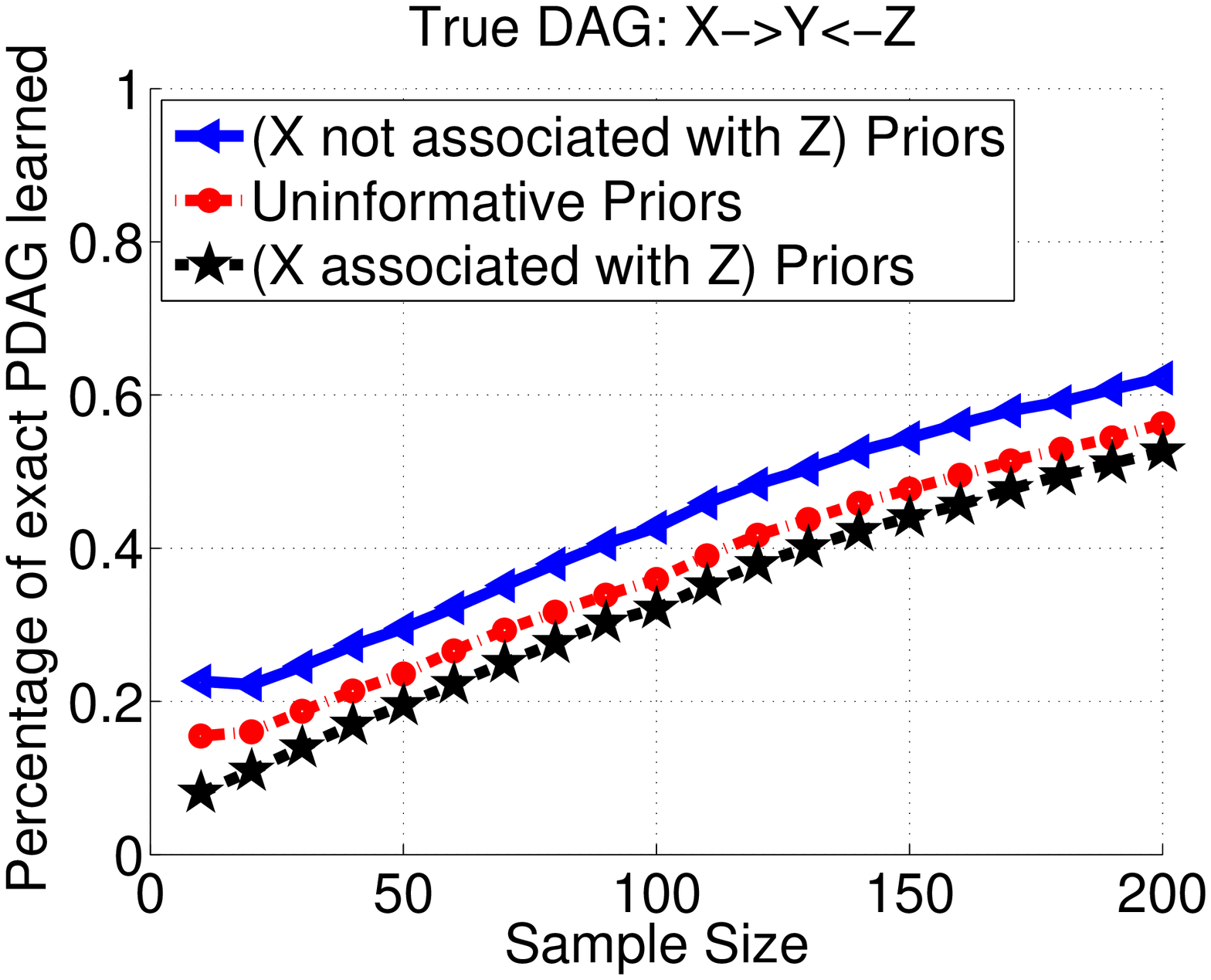}
}
\subfloat[][]{\label{fig:alarm}
\includegraphics[width=0.49\columnwidth]{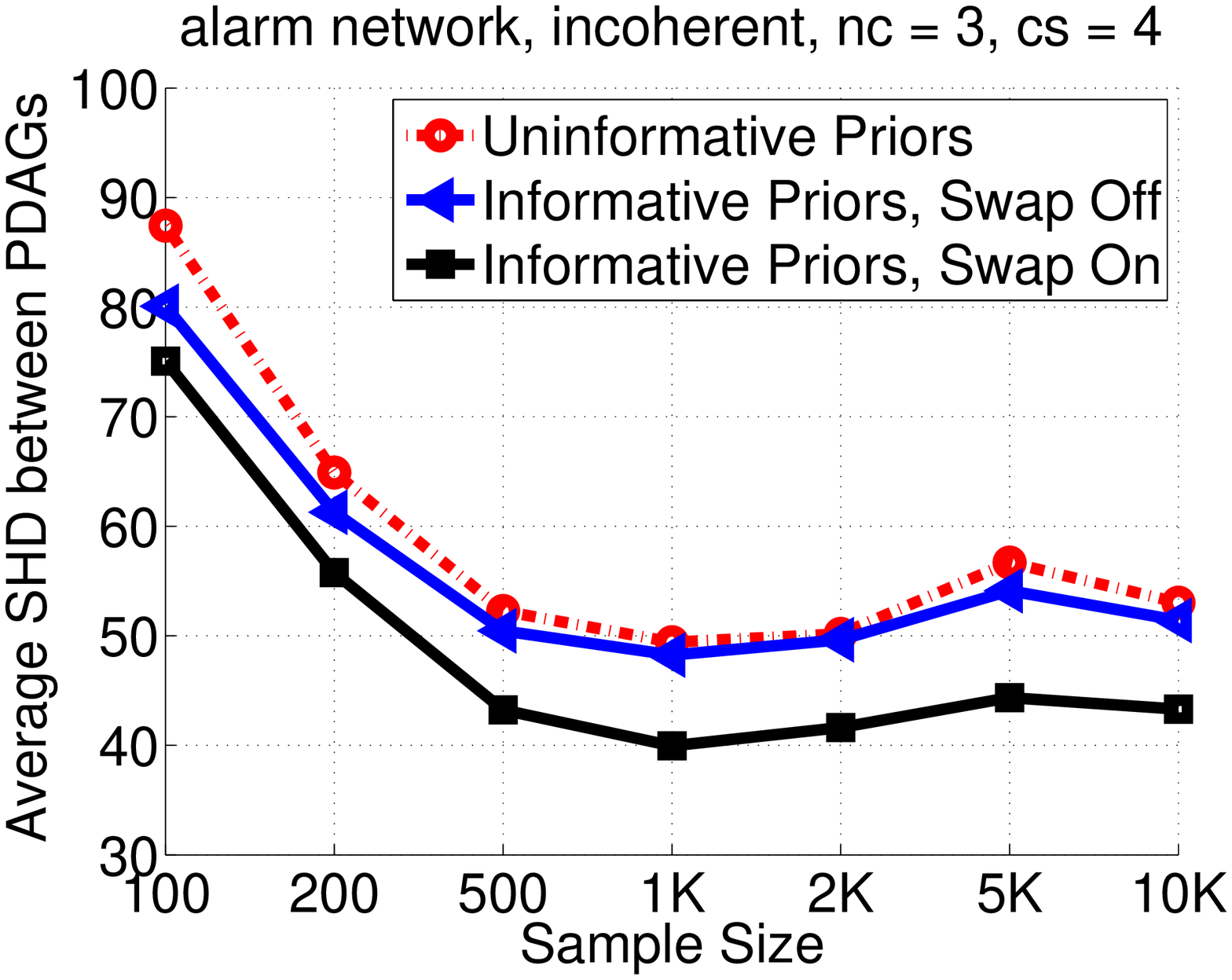}
}
\subfloat[][]{\label{fig:insurance}
\includegraphics[width=0.49\columnwidth]{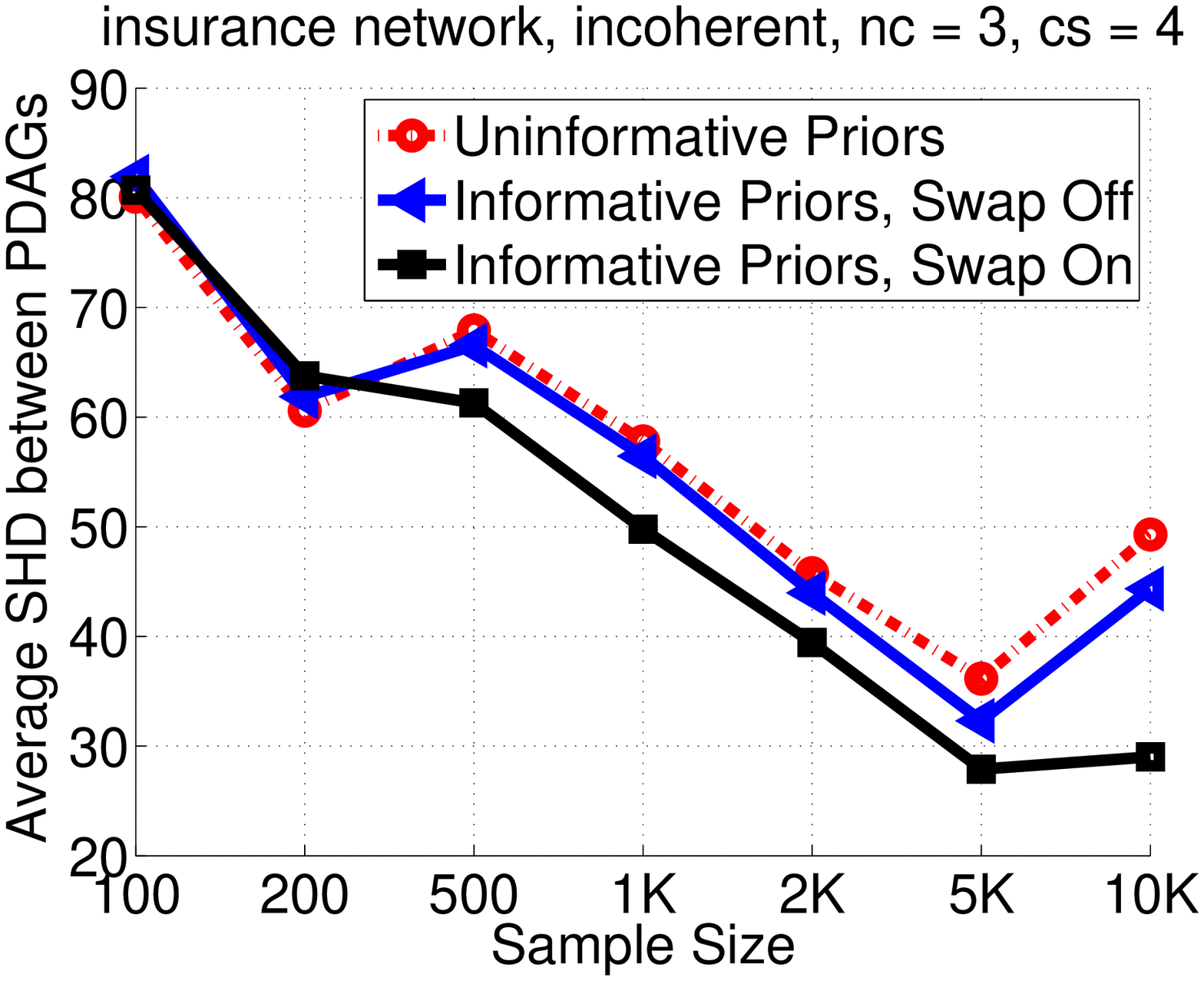}
}
\caption{
(a) Learning the orientations and the skeleton is facilitated by causal prior knowledge. (b) Learning the graph is facilitated by correct associative prior knowledge and hindered by incorrect priors. (c-d) Learning the ALARM and INSURANCE networks. The average Structural Hamming Distance (SHD) is shown with increasing sample size, for component size (cs) 4 and number of components (nc) set to 3, and incoherent beliefs. Using path beliefs, especially combined with the swap-equivalent operator, produces better networks on average.}
\vspace{-0.2cm}
\end{figure*}%

\textbf{Employing Causal Knowledge}. 
We consider the graph $X \rightarrow Y \rightarrow Z$. 
We use the path belief $P(X \rightpath Z) = 0.9$ and distribute the remaining 0.1 mass of probability to the remaining values of $r_{XZ}$ proportional to the values that correspond to a uniform prior. 
We repeat the following experiment $10000$ times: (a) we randomly select the number of states for each variable to be either 3 or 4, (b) we sample the cpts for each variable from the gamma distribution $\Gamma(k,\theta)$, with shape parameter $k$ set to $0.5$ and scale parameter $\theta$ set to $1$, (c) we sample a dataset of size $200$, (d) we increase the samples of the dataset provided to the scoring method from 10 to 200 with step size of 10, (e) we identify the highest scoring network out of all 25 possible DAGs using informative and uninformative priors and the BDeu score with Equivalent Sample Size (ESS) set to 1.

{\bf Results:} Figure \ref{fig:exactNcollider} plots the percentage of the time the PDAG $X - Y - Z$ of the true network was found exactly, with and without informative priors. 
First notice, that when the true PDAG is found, the edges are also {\em always oriented correctly since the true network has a higher prior than any other Markov-equivalent graph}. 
Perhaps more surprising though, notice that the informative priors also {\em improve the learning of the skeleton.} 
The belief $X \rightpath Z$ tends to add a path from $X$ to $Z$. 
The associations $X-Y$ and $Y-Z$ are always higher than or equal to the association between $X-Z$ \citep{Cover2006}. 
Thus, {\em it is the correct path $X - Y - Z$ that tends to be induced, rather than any other network with a path $X \rightpath Z$}.

\textbf{Employing Associative Knowledge}. 
We run a similar proof-of-concept experiment where the true network is a single collider $X \rightarrow Y \leftarrow Z$. 
We use the same settings as before for three cases: correct associative priors $P(X \nopath Z) = 0.9$, uniform priors, and incorrect associative priors $P(X \text{ associated with } Z) = 0.9$.

{\bf Results:} 
The results are shown in Figure \ref{fig:exactcollider}. 
As expected, {\em correct prior beliefs clearly improve the chances of identifying the true PDAG; the effect is exactly the opposite when misleading, incorrect beliefs are provided to the algorithm}. 
Of course, asymptotically any non-zero priors play no role.

\textbf{Learning Larger Networks}. 
To generate path beliefs we use three parameters: the number of independent components $nc$, the number of nodes appearing in an independent component $cs$, and whether we want them to be coherent or incoherent.
Path variables were generated as follows: for given $cs$ and $nc$, we randomly pick $nc$ non-overlapping sets, each containing $cs$ nodes of the network, and consider all possible pairs between them as path variables, resulting in a total of $nc \cdot cs \cdot (cs - 1) / 2$ path variables.
This is done in order to be able to consider large sets of path variables.
Then, we randomly assign a probability $p \in [0.5,0.99]$ to the true value of each path variable, and split the remaining $1 - p$ mass probability in an uninformative way to the remaining values.
This process is repeated for each independent component until it is coherent or incoherent, depending on the input parameter.
To estimate $U$ we sampled $S = 10^6$ DAGs and $l$ was set to the machine epsilon.
We used the ALARM \citep{Beinlich.etal.1989} and the INSURANCE \citep{Insurance} networks to evaluate our methods.
We employed Greedy Search with the BDeu metric and ESS=1. 
We run the method starting from the empty graph with uninformative and informative priors, as well as with and without the swap-equivalent-operator in the case of informative priors.
Finally, we compute the Structural Hamming Distance \citep{Tsamardinos.etal.2006} from the PDAG of the true network. 
We used the PDAG to avoid introducing an unfair advantage for our methods; all methods may find Markov equivalent DAGs, but the ones using path beliefs may find more correctly oriented edges.
The sample size was varied within $\{100, 200, 500, 1000, 2000, 5000, 10000\}$.
The path belief parameters were varied within $\{1,2,3,4,5\}$ and for $nc$ and $cs$ respectively, for both the coherent and incoherent cases.
The experiment was repeated 100 times, for randomly sampled datasets and path beliefs, with all combinations of input parameters.

\textbf{Results:} 
Due to space limitations we report only the results for incoherent path beliefs, with $nc = 3$ and $sc = 4$ (18 beliefs).
The results were similar for both, coherent and incoherent priors.
Also, with smaller (larger) $nc$ and $sc$, the difference between the uninformative and informative methods was smaller (larger).

The results are shown in Figures \ref{fig:alarm} and \ref{fig:insurance}. 
\textit{In all cases, the SHD is smaller with the informative priors than with uninformative priors.} 
For the ALARM network, notice that the SHD difference between the uninformative method and the informative method without the operator decreases as sample size increases.
The reason is that, as sample size increases, the data score becomes more important and the prior score tends to be ignored; it usually is considered only close to local maxima, where only small improvements in the data score can be made.
If however the swap-equivalent operator is used, this does not happen, as it tries to maintain a high prior score during the whole search.
Finally, notice the counter-intuitive behavior of increasing SHD with increasing sample size in Figure \ref{fig:insurance} for 10K samples.
Anecdotal experiments suggest that the value of the ESS parameter is the reason for that behavior.
However, when the swap-equivalent operator is used, this phenomenon is almost nonexistent. 
\section{CONCLUSIONS}

We present a method for computing informative priors given a set of causal and associative beliefs on pairs of variables, as well as a novel search-operator to take advantage of them. The priors can then be employed by any search-and-score learning algorithm. 
The method, for the first time, addresses the issues of incoherent and possibly dependent priors. Providing correct priors about pairwise causal or associative relations improves learning both in terms of identifying the orientation of the edges (for causal priors), but also in terms of identifying the skeleton of the network.

There are numerous issues to still address regarding both the method and the general problem. 
The algorithm has exponential worst-case time complexity, thus more efficient algorithms are desirable.
Closed-form solutions for computing the number of graphs given path constraints are also desirable. 
Finally, including other types of prior knowledge, as well as incorporating the strength of the causal effects or associations and other prior knowledge characteristics is an interesting future direction to pursue.



\subsubsection*{Acknowledgements}
This work was partially funded by EU FP7 No 306000 STATegra, EU FP7  248590 REACTION, and Greek GSRT Thalis 20332 Symbiomics. We would like to thank the anonymous reviewers and Prof. G. F. Cooper for suggestions.

%
%
%
%
%

\bibliographystyle{abbrvnat}
\renewcommand\refname{\subsubsection*{References}}
\bibliography{uai2013}

%
%

\end{document}